\documentclass[11pt]{article}
\usepackage[margin=1in]{geometry}
\usepackage{amsmath,amsfonts}
\usepackage{algorithmic}
\usepackage{algorithm}
\usepackage{array}
\usepackage[caption=false,font=normalsize,labelfont=sf,textfont=sf]{subfig}
\usepackage{textcomp}
\usepackage{url}
\usepackage{verbatim}
\usepackage{graphicx}
\usepackage{cite}
\usepackage[table]{xcolor}
\usepackage{booktabs}
\usepackage{multirow}
\usepackage{caption}
\usepackage[hidelinks]{hyperref}
\newlength{\ieeecolwidth}
\begin{document}

\title{Shortest-Path Decomposition for Foliage-Robust 3D Tree Modeling and Above-Ground Biomass Estimation from Point Clouds}

\author{Di Wang\thanks{Corresponding author. Email diwang@xjtu.edu.cn.} and Shi Li\thanks{All authors are with the School of Software Engineering; the State Key Laboratory of Human-Machine Hybrid Augmented Intelligence; and the Shaanxi Joint (Key) Laboratory for Artificial Intelligence, Xi'an Jiaotong University, Xi'an 710049, China.}}

\maketitle

\begin{abstract}
Estimating above-ground biomass (AGB) from terrestrial laser scanning (TLS) via quantitative structural models (QSMs) is highly accurate under leaf-off conditions, yet major QSM paradigms degrade sharply when foliage is present. The standard remedy, leaf--wood separation, introduces its own errors and complexity. We present topology-driven foliage suppression that replaces explicit leaf--wood classification inside the reconstruction pipeline. On the shortest-path tree of a point-cloud graph, the number of root-to-node paths traversing a node equals the size of its rooted subtree, so path traversal frequency is an exact topological descriptor of branching hierarchy. Three decompositions of the path structure, plus an adaptive frequency threshold at the elbow of frequency growth along each branch, recover the branching hierarchy from a single Dijkstra run. We evaluate it on 90 trees under exclusively leaf-on conditions across tropical TLS and temperate UAV laser scanning (ULS) platforms. On dense tropical TLS it achieves a mean absolute percentage deviation (MAPD) of 17.2\% with RMSE 2.37~Mg (RMSE\% 34.8\%) and an $R^2$ of 0.956, outperforming the best separation-assisted TreeQSM pipeline (21.2\%). Structural analysis confirms that foliage introduces 2.6~cm of mean geometric error (3.8~cm RMS) and preserves the hierarchical volume distribution. On low-density ULS data, MAPD reaches 6.4\% with field-measured diameter at breast height (DBH) and 32.3\% with DBH inferred allometrically from tree height. All AGB experiments use identical parameter settings. Directly recovering branching hierarchy from shortest-path structure removes a barrier to operational QSM-based forest inventory under leaf-on conditions.
\end{abstract}


\section{Introduction}
\label{sec:intro}

Accurate estimation of forest above-ground biomass (AGB) is essential for carbon accounting, climate policy, and ecosystem monitoring~\cite{Goetz2009,duncanson2019importance}. Laser scanning, whether terrestrial (TLS), mobile (MLS), or UAV-based (ULS), captures three-dimensional tree structure at millimeter to centimeter resolution and supports non-destructive biomass estimation through quantitative structural models (QSMs). A QSM reconstructs tree branching architecture, namely how branches connect (topology) and how thick they are (geometry). Wood volume is computed from this reconstruction; multiplying by species-specific wood density yields AGB estimates more accurate than traditional allometric equations, especially for large trees~\cite{demol2022estimating,ali2025hybrid,gonzalez2018estimation}. These estimates increasingly serve as ground-reference data for calibrating airborne and spaceborne biomass products~\cite{labriere2023toward,brede2022nondestructive}.

Two families of QSM methods dominate the literature. Cylinder-fitting methods, led by TreeQSM~\cite{Raumonen2013}, SimpleTree~\cite{Hackenberg2015}, and PypeTree~\cite{Delagrange2014}, partition the point cloud into segments and fit cylinders directly to local points. Skeleton-based methods, including AdTree~\cite{Du2019}, AdQSM~\cite{Fan2020}, and L1-Tree~\cite{feng2024l1}, abstract the point cloud to a graph and extract a branching centerline. Both families, however, share a critical vulnerability, foliage. Under leaf-on conditions, cylinder fitting inflates volume by treating dense leaf clusters as woody material, a well-documented failure mode in the QSM literature~\cite{chen2025impact}, while skeleton-based methods generate spurious branches because the underlying graph must connect every leaf point.

The standard remedy is a leaf--wood separation preprocessing step, applied before the QSM pipeline~\cite{wang2020lewos}. However, leaf--wood separation introduces its own problems. A recent benchmark of 11 published algorithms found that while the best methods achieved relative AGB differences of $\pm$2--3\%, most produced systematic underestimation of 2--46\%, with fine peripheral branches easily misclassified as leaves and removed~\cite{chen2025impact}. Critically, the classifier and the reconstructor are designed and optimized independently. Classification errors therefore propagate into the downstream QSM with no guarantee of bounded or predictable effect, an architectural fragility that no improvement in classifier accuracy alone can resolve, because the interface between the two stages is untested by either.

The bottleneck is therefore architectural rather than algorithmic. The prevailing assumption, embedded in every QSM pipeline that depends on preprocessing, is that leaves and wood must be separated before either can be understood. An alternative is a reconstruction principle that bypasses this assumption entirely, relying on a signal that distinguishes trunk from periphery without classifying individual points, a signal that is a property of the whole tree. Such a signal exists in the shortest-path structure of a point-cloud graph. When all points are connected into a graph and shortest paths are computed from the tree base to every other point, the branching architecture is encoded in the path traversal pattern. A shortest path from the base to a canopy point must traverse the trunk and major branches. Arterial woody edges are therefore traversed by tens of thousands of paths, while foliage points form peripheral clusters reached by few paths. Because this frequency gap arises from connectivity structure, it persists regardless of leaf shape, point density, or sensor platform. Vicari et al.~\cite{vicari2019leaf} first observed that path traversal frequency reflects hierarchical position and used it for leaf--wood classification.

To exploit this topological signal for reconstruction, we develop a method built on three decompositions of the shortest-path structure, all derived from a single Dijkstra run. First, path frequency identifies arterial nodes by the number of paths traversing them. Second, a farthest-tip assignment partitions the point cloud into branch-level clusters. Third, a reverse-distance metric segments each branch along its length. An adaptive frequency threshold, derived from the elbow point of frequency growth along each branch, suppresses leaf-level nodes without per-tree tuning. Together, these decompositions recover the branching hierarchy and form a QSM from which AGB is computed, with topology-driven foliage suppression, no geometric feature-based classification, and no per-dataset parameter adjustment.

We evaluate the method on 90 trees under exclusively leaf-on conditions across tropical TLS and temperate ULS platforms, comparing against TreeQSM, AdQSM, and their combinations with leading leaf--wood separation algorithms. All AGB experiments use identical parameter settings.

The contributions of this paper are as follows.
\begin{enumerate}
\item We reformulate QSM reconstruction as a topology recovery problem rather than a semantic segmentation problem. In this framework, the branching hierarchy is extracted directly from graph connectivity, and geometric modeling is performed on the recovered topology rather than on raw point clusters.
\item We establish the theoretical foundation for topology-guided skeleton extraction, showing that on a rooted tree the number of root-to-node paths traversing a node equals the size of its rooted subtree and grows monotonically toward the root. This makes path traversal frequency an exact topological descriptor and underpins foliage suppression without geometric feature-based classification.
\item We develop a topology-first reconstruction pipeline based on three decompositions of the shortest-path structure, all derived from a single Dijkstra run. An adaptive elbow threshold, which detects relative frequency jumps at branch junctions, replaces manual parameter specification.
\item We demonstrate that topology-first reconstruction transfers across sensor platforms (TLS and ULS) and biomes (tropical and temperate) without parameter adjustment. Structural consistency analysis on 61 tropical trees confirms that foliage introduces 2.6~cm of mean geometric error (3.8~cm RMS) relative to leaf-off conditions, and that the hierarchical volume distribution is preserved.
\end{enumerate}

\section{Related Work}
\label{sec:related}

\subsection{QSM Paradigms and the Foliage Bottleneck}

Cylinder-fitting methods, led by TreeQSM~\cite{Raumonen2013}, SimpleTree~\cite{Hackenberg2015}, and PypeTree~\cite{Delagrange2014}, partition the point cloud into segments and fit cylinders directly to local points. This family underpins non-destructive AGB estimation from TLS~\cite{calders2015nondestructive}, with extraction pipelines such as treeseg~\cite{burt2019treeseg} supplying the individual-tree point sets that QSM pipelines consume and branch-architecture studies relying on cylinder models in tropical forests~\cite{lau2018branch}. They achieve state-of-the-art accuracy on clean TLS input~\cite{ali2025hybrid} but are fragile under leaf-on conditions, where dense foliage clusters are fitted as woody cylinders, inflating volume. TreeQSM's AGB error increases sharply on unfiltered leaf-on data~\cite{chen2025impact}. The foliage bottleneck is well documented, and reviews of TLS in forest inventories~\cite{liang2016tlsinventory} and in forest ecology~\cite{calders2020expanding} consistently identify leaf--wood separation as a central obstacle to QSM-based AGB estimation~\cite{disney2018weighing}. Skeleton-based methods, including AdTree~\cite{Du2019}, AdQSM~\cite{Fan2020}, and L1-Tree~\cite{feng2024l1}, abstract the point cloud to a graph and extract a branching centerline. They gain some occlusion tolerance but remain vulnerable to foliage because the graph must connect every input point, producing spurious branches through leaf clusters. Beyond these two mainstream families, several general-purpose skeletonization techniques extract centerlines through graph contraction~\cite{jiang2021skeleton}, graph skeletonization~\cite{bucksch2008campino}, or L1-median optimization~\cite{mei20173d}, and tools such as PC-Skeletor~\cite{meyer2023pcskeltor} and CherryPicker~\cite{meyer2023cherrypicker} provide point-cloud skeletonization libraries. These methods, however, rely on local geometric operations and produce skeletons sensitive to foliage-induced surface perturbations. An emerging third paradigm uses deep learning. TreePartNet~\cite{Liu2021}, DeepTree~\cite{zhou2023}, TreeStructor~\cite{Zhou2025}, Smart-Tree~\cite{dobbs2023smarttree}, and SmartQSM~\cite{lin2026smartqsm} predict medial-axis or component properties from point clouds. They rely on labelled training data, which public datasets such as~\cite{biodiv2025} now supply, yet the synthetic training data typically used for these methods may not match real forest conditions, and generalization to unseen species and sensor configurations remains an open question. The learned mapping from point cloud to skeleton is also opaque; when a prediction fails, the failure mode is difficult to diagnose or correct.

Among graph-based QSM pipelines, RayExtract~\cite{devereux2026rayextract} reconstructs woody volume from a shortest-path forest and processes leaf-on point clouds without explicit leaf--wood separation, reporting accurate total volume against destructive harvest references. Its leaf tolerance is geometric, however, relying on segment crop length, robust averaging of fitted radii, and morphological taper priors, and branch-level structural validation is deferred to future work.

Our method belongs to the skeleton-based family but adopts a topology-first approach, in which the branching hierarchy is recovered from graph connectivity before geometric modeling, rather than as a by-product of cylinder fitting or point classification. It requires no training data, unlike learning-based methods, and addresses foliage at the mechanism level, removing leaf-level nodes before branch radii are estimated.

\subsection{Leaf--Wood Separation and Its Limitations}

Leaf--wood separation has been extensively studied as a preprocessing step for QSM. Early approaches used radiometric features (exploiting the difference in laser return intensity between leaf and wood surfaces) or geometric features such as local curvature, dimensionality, and normal orientation. Density-based clustering (DBSCAN) was applied by~\cite{FERRARA2018434} to group points by spatial proximity before classification. Methods based on geometric features alone include LeWoS~\cite{wang2020lewos}, which uses recursive geometric segmentation with a classification threshold derived from the point distribution. GBS~\cite{Tian2022} exploits shortest-path connectivity patterns, treating leaf points as graph leaves that are pruned by analyzing edge betweenness. More recently, deep-learning methods have achieved up to 92.2\% mean IoU on pointwise classification~\cite{vandenbroeck2025pointwise}, though their performance depends on training data representativeness.

A comprehensive benchmark of 11 algorithms by~\cite{chen2025impact}, spanning radiometric, geometric, graph-based, and learning-based approaches, revealed systematic AGB underestimation of 2--46\% depending on forest type and algorithm, with the best methods achieving relative AGB differences of $\pm$2--3\%. The primary source of error across all algorithm categories is the misclassification of fine peripheral branches as leaves, particularly at branch extremities where the local geometry of a thin branch is nearly indistinguishable from that of a leaf. The deeper architectural issue is that the classifier and the reconstructor are designed and optimized independently, so classification errors propagate into the downstream QSM with no guarantee of bounded or predictable effect, and the interaction between the two stages at the point-cloud level is untested by either. Our method avoids this two-stage architecture. Foliage suppression emerges from hierarchical path-frequency decomposition.

\subsection{Path-Frequency Methods for Tree Analysis}

The earliest use of shortest-path structure for tree modeling was by Livny et al.~\cite{livny2010automatic}, who reconstructed skeletal structures from point clouds via global optimization on a shortest-path tree. Graph pathing, introduced by~\cite{WANG202167}, represents the point cloud as a KNN--Delaunay hybrid graph and extracts individual trees by routing each point to its lowest reachable node. Vicari et al.~\cite{vicari2019leaf} were the first to observe that shortest-path traversal frequency reflects hierarchical position within a tree and used it as a feature for leaf--wood classification. Li et al.~\cite{li2024efficient} applied graph-based shortest-path methods to fruit-tree skeleton extraction. These prior works established path frequency as a useful signal but used it either for classification or for skeleton extraction in controlled orchard environments. We extend this line of work by using path-frequency decomposition as the sole skeleton-extraction mechanism in a full QSM pipeline operating on natural forest trees. This topology-first approach replaces both geometric feature-based classification and explicit leaf--wood separation.

\subsection{Cross-Platform QSM from TLS to ULS}

Most QSM methods are validated exclusively on TLS data. Brede et al.~\cite{brede2019nondestructive,brede2022nondestructive} demonstrated that ULS can support population-level AGB estimation when combined with TLS-QSM reference data. Fekry et al.~\cite{fekry2022} fused ground-based and UAV-LiDAR data for QSM-based tree parameter retrieval in subtropical forests, reporting that fusion improved DBH estimation accuracy compared to ULS alone. Terryn et al.~\cite{TERRYN2022} quantified tropical forest structure through TLS and ULS fusion in Australian rainforests, highlighting the complementarity of the two platforms. Gan et al.~\cite{gan2024nondestructive} reported that both TreeQSM and AdQSM struggled with DBH estimation from aerial platforms. Reckziegel et al.~\cite{reckziegel2025assessing} found ULS and TLS comparable in tropical savanna. Luo et al.~\cite{luo2026inceptionformer} proposed deep-learning point cloud completion for dense forests. Across these studies, ULS-based QSM requires fusion with ground data or TLS calibration. Our method achieves parameter consistency across platforms by relying on graph topology rather than point density.

\begin{table}[t!]
\centering
\caption{Summary of Datasets}
\footnotesize
\begin{tabular}{l c c c c}
\toprule
Dataset & Platform & Leaf Condition & Biome  & Trees \\
\midrule
Cameroon & TLS & Leaf-off + Leaf-on & Tropical  & 61 \\
Germany  & ULS & Leaf-on & Temperate & 29 \\
\bottomrule
\end{tabular}
\label{tab:datasets}
\end{table}

\section{Datasets}
\label{sec:datasets}

We evaluate on two datasets spanning tropical and temperate biomes, TLS and ULS platforms, and both leaf-on and leaf-off conditions. Table~\ref{tab:datasets} summarizes their key characteristics. Destructive AGB measurements are available for Cameroon; for the ULS dataset, co-registered TLS-derived AGB serves as reference.

\subsection{Cameroon Tropical Forest TLS}

\begin{figure}[t]
    \centering
    \includegraphics[width=\ieeecolwidth]{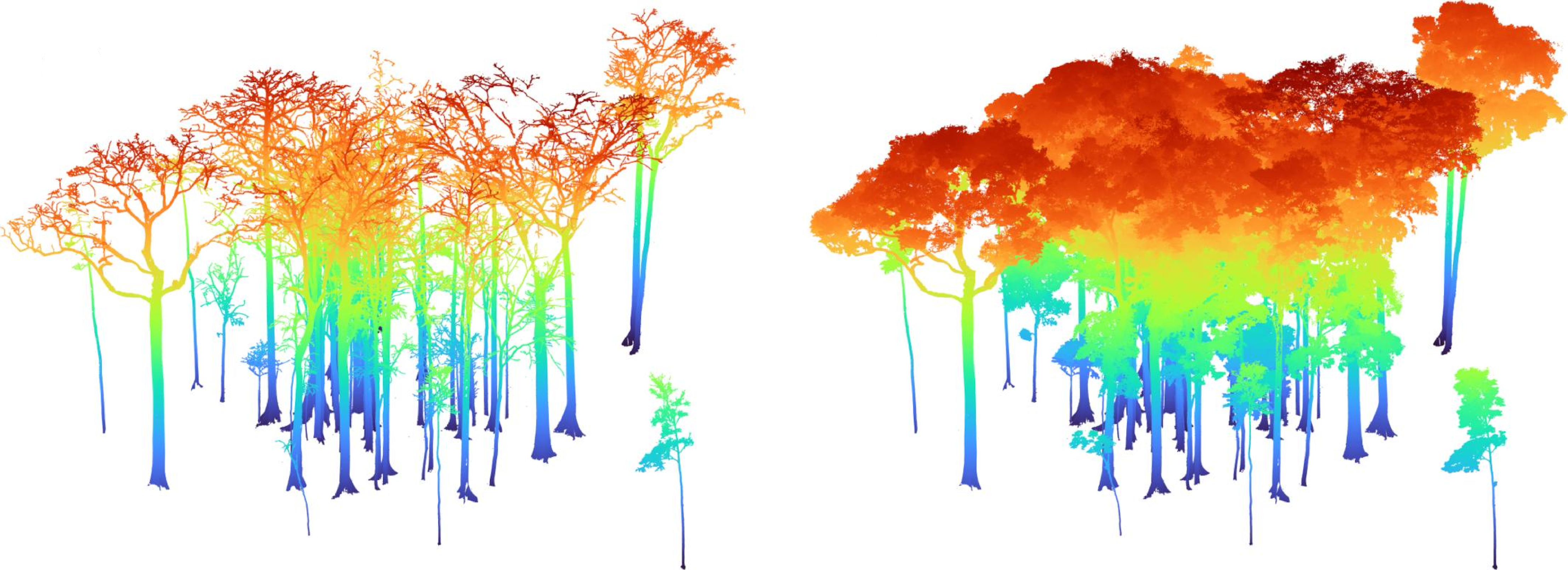}
    \caption{Point clouds of the 61 Cameroon trees. Left, leaf-off. Right, leaf-on.}
    \label{fig:momovis}
\end{figure}

\begin{figure}[t!]
    \centering
    \includegraphics[width=0.6\ieeecolwidth]{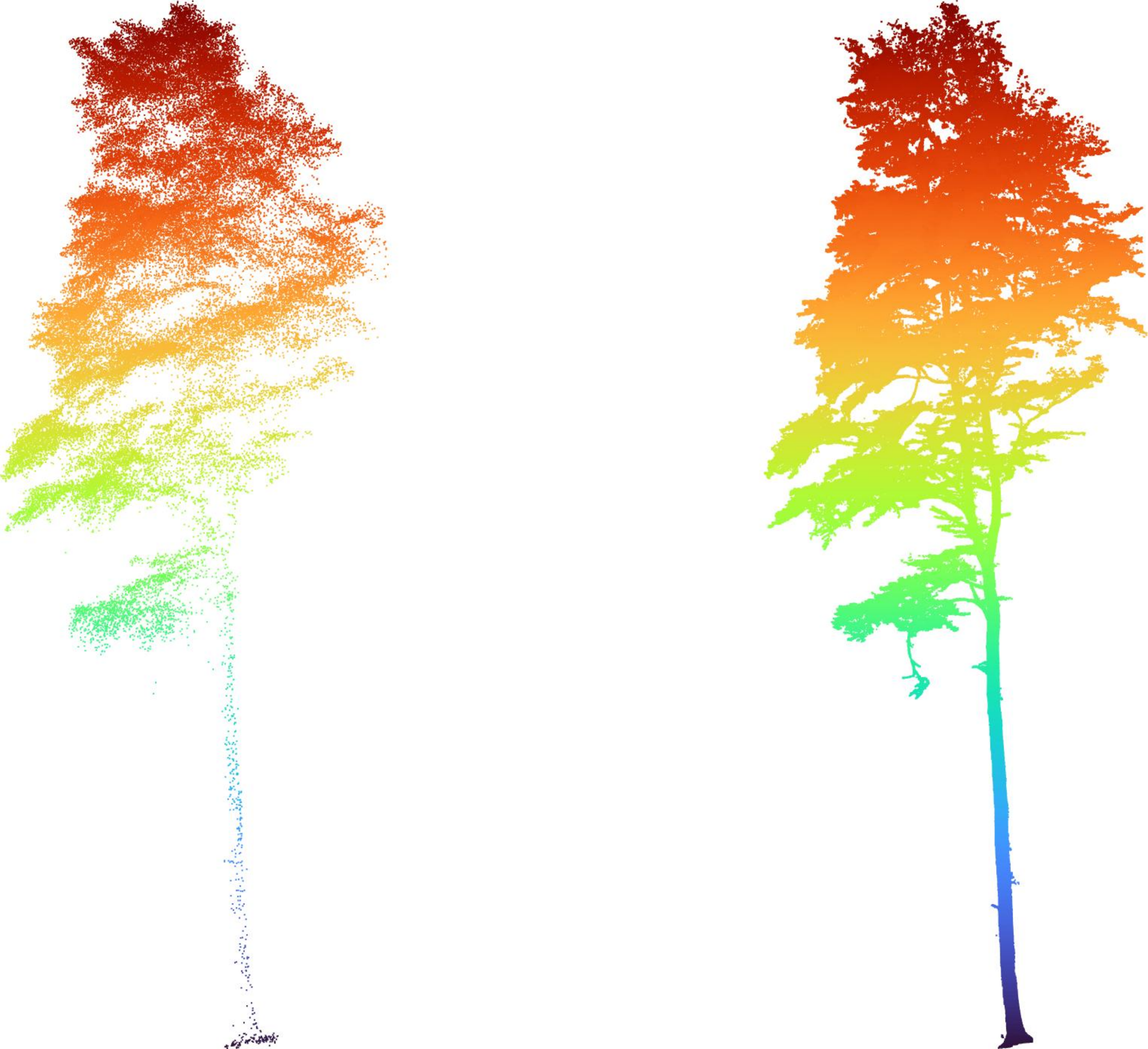}
    \caption{Paired ULS (left) and TLS (right) point clouds for the same tree.}
    \label{fig:ulsvis}
\end{figure}

{\sloppy\par This dataset contains 61 tropical trees from semi-deciduous forests in eastern Cameroon ($4^\circ 02' 20.77''$N, $14^\circ 55' 49.15''$E), near Lob\'{e}k\'{e} National Park~\cite{momomee2018}. The trees represent 15 species, with a mean DBH of 58.4~cm ($\pm$41.3) and a mean height of 33.7~m ($\pm$12.4). The maximum stem diameter reaches 186.6~cm. All 61 trees were destructively harvested, providing ground-truth AGB measurements and a unique benchmark for evaluating biomass estimation in structurally complex tropical trees.\par}

TLS data were acquired with a Leica C10 ScanStation (Leica Geosystems), a time-of-flight laser scanner operating at a 532~nm wavelength. Each tree was scanned from at least three positions to ensure comprehensive coverage. Point clouds were acquired under both leaf-off and leaf-on conditions for the same 61 trees (Fig.~\ref{fig:momovis}). In the leaf-off data, leaf points were manually removed by the original authors to facilitate accurate 3D model reconstruction~\cite{momomee2018}. The leaf-off dataset serves as a reference ceiling for evaluating the best-case performance of cylinder-fitting QSMs (Section~\ref{sec:cameroon}). The leaf-on dataset, which retains all foliage points, is the primary test bed for evaluating our method's foliage robustness. All methods are evaluated on the same 61 trees, enabling direct comparison across leaf conditions.

\subsection{Germany Temperate Mixed Forest ULS}

The ULS dataset covers a temperate mixed forest in the Bretten municipal forest, Baden-W\"{u}rttemberg, Germany~\cite{essd-14-2989-2022}. The site is a hilly landscape with multi-layered, mixed-species stands characterized by dense canopy cover. Dominant species include \textit{Fagus sylvatica}, \textit{Pseudotsuga menziesii}, \textit{Picea abies}, and \textit{Quercus} spp.

ULS data were acquired with a RIEGL miniVUX-1UAV sensor mounted on a DJI Matrice 600 Pro UAV during multiple flight campaigns between 2019 and 2020. A double-grid flight pattern ensured full coverage. Point clouds were post-processed with RiPROCESS software and aligned with GNSS reference data; outliers and off-nadir noise were filtered. Co-registered TLS data were collected with a RIEGL VZ-400 scanner in 2019 using 5--8 scan positions per plot, registered via reflective cylindrical targets and multi-scan adjustment, then aligned with ULS data using iterative closest point (ICP) algorithms.

The dataset provides manually segmented paired ULS and TLS point clouds for 29 individual trees (Fig.~\ref{fig:ulsvis}). ULS point densities are substantially lower than TLS, and lower-trunk coverage is incomplete due to canopy occlusion. Destructive AGB measurements are unavailable; TLS-derived AGB (computed from the co-registered TLS data with measured DBH) serves as reference, a practice established in prior ULS-based AGB studies~\cite{brede2019nondestructive,brede2022nondestructive}. This dataset tests the method under the most challenging conditions, namely low density, partial coverage, and leaf-on foliage simultaneously.

\section{Method}
\label{sec:method}

\subsection{Topological Basis of Path Frequency}
\label{sec:theory}

The method rests on a property of rooted tree graphs. Every node is connected to the root by a unique simple path, so the number of root-to-node paths traversing a node directly encodes its topological position.

Consider a rooted tree $T = (V_T, E_T)$ with node set $V_T$, edge set $E_T$, and root $r \in V_T$. For each node $u \in V_T$, there exists a unique simple path $P(r, u)$ from the root to $u$. The \emph{path frequency} of node $v$ is
\begin{equation}
F(v) = \sum_{u \in V_T} \mathbf{1}\bigl(v \in P(r, u)\bigr),
\label{eq:F_theory}
\end{equation}
where $\mathbf{1}(\cdot)$ is the indicator function. $F(v)$ counts how many root-to-node paths traverse $v$.

\begin{quote}
\textbf{Theorem 1.} Let $\operatorname{Sub}(v) = \{u \in V_T \mid v \in P(r, u)\}$ be the set of nodes whose root path passes through $v$, i.e., the subtree rooted at $v$. Then $F(v) = |\operatorname{Sub}(v)|$.
\end{quote}

\begin{quote}
\textit{Proof.} Since $T$ is a tree, $P(r, u)$ is unique for every $u$. If $u \in \operatorname{Sub}(v)$, then by definition $v \in P(r, u)$ and $u$ contributes $1$ to $F(v)$. If $u \notin \operatorname{Sub}(v)$, then $v \notin P(r, u)$ and the contribution is $0$. Summing over all $u \in V_T$ yields $F(v) = |\operatorname{Sub}(v)|$.
\end{quote}

Theorem~1 states that on a rooted tree, path frequency equals the number of descendant nodes in the subtree rooted at each node, a definitional property rather than a statistical approximation.

\begin{quote}
\textbf{Corollary 1.} Let $u$ be a child of $v$ in $T$. Then $\operatorname{Sub}(u) \subset \operatorname{Sub}(v)$, implying $F(v) \geq F(u)$. Path frequency is monotonically non-decreasing along any rootward path.
\end{quote}

Corollary~1 provides a natural hierarchical ordering, in which frequency increases from leaf tips toward the root without any semantic classification of points. The behavior at branch junctions follows directly. Along a single branch, subtree size changes incrementally, producing a gradual frequency increase. At a junction where $k$ sibling subtrees merge, the frequency jumps by $\sum_{i=1}^{k} |\operatorname{Sub}(u_i)|$. These discontinuities mark topological transitions (where distinct branches join) rather than artifacts of local point geometry. The adaptive threshold (Section~\ref{sec:adaptive}) is designed to detect these frequency jumps.

In our pipeline, the proximity graph $G = (V, E, W)$ constructed from the point cloud (Section~\ref{sec:graph}) is not strictly a tree; it is an undirected graph that may contain redundant edges, oriented into two directed arcs per edge for extraction (Section~\ref{sec:graph}). A directed Dijkstra run~\cite{dijkstra1959} extracts a shortest-path tree $T \subseteq G$ rooted at the base node $v_r$, and Theorem~1 applies to this implicit tree. The propagation step (Eq.~\ref{eq:Fi_correct}) restores the monotonic ordering guaranteed by Corollary~1 in the presence of graph construction noise. Theorem~1 also explains why path frequency is robust to local perturbations, since spurious edges affect only a small fraction of shortest paths, leaving the global topological signal intact.

The practical implication for skeleton extraction follows directly. Nodes on the main stem and primary branches support substantially larger descendant subtrees than terminal twigs. Path frequency therefore separates trunk, primary branches, and fine branches according to their hierarchical position rather than their local geometry. The adaptive threshold (Section~\ref{sec:adaptive}) detects transitions between these hierarchical levels by identifying relative frequency jumps at branch junctions, where multiple subtrees merge. Because the frequency ratio across a junction reflects the relative sizes of the merging subtrees, the threshold criterion is scale-invariant, depending on the topological hierarchy encoded in frequency ratios, not on absolute tree size or point density.

\begin{algorithm}[t]
\caption{Skeleton Extraction via Shortest-Path Decomposition}
\label{alg:pipeline}
\begin{algorithmic}
\REQUIRE Point cloud $P$, parameters $k$, $a$, weight $\beta$
\ENSURE Skeleton nodes $S$, directed graph $G_{\text{skel}}$
\STATE $G \gets \textsc{BuildGraph}(P, k)$ \hfill // KNN + Delaunay, prune, connect
\STATE $\{\mathcal{P}\} \gets \textsc{DirectedDijkstra}(G, v_r, \beta)$ \hfill // single run, all-to-root
\STATE $D_i \gets \textsc{EuclideanPathLength}(\mathcal{P})$ \hfill // root distances in meters
\STATE $\mathcal{F}' \gets \textsc{PropagateCorrect}(\mathcal{P})$ \hfill // Eq.~4, hierarchical decomp.
\STATE $\mathcal{T}' \gets \textsc{PropagateCorrect}(\mathcal{P}, D)$ \hfill // Eq.~5, branch decomp.
\STATE $\mathcal{B} \gets \textsc{PartitionByTip}(P, \mathcal{T}')$ \hfill // one cluster per farthest tip
\FORALL{cluster $B \in \mathcal{B}$}
\STATE $\mathbf{d}_s \gets \textsc{ExpBins}(D_T, a)$ \hfill // Eqs.~7--8, spatial decomp.
\STATE $S \gets S \cup \textsc{HighestFreqNode}(B, \mathcal{F}')$ \hfill // skeleton node per bin
\ENDFOR
\STATE $\mathcal{F}_{\text{adapt}} \gets \textsc{ElbowThreshold}(S, \mathcal{F}')$ \hfill // Eq.~9, adaptive
\STATE $S \gets S[\mathcal{F}' \ge \mathcal{F}_{\text{adapt}}]$ \hfill // suppress leaf-level nodes
\STATE $G_{\text{skel}} \gets \textsc{Reconnect}(S, \mathcal{P})$ \hfill // edges from path abstraction
\RETURN $S$, $G_{\text{skel}}$
\end{algorithmic}
\end{algorithm}

\begin{figure*}[tp]
    \centering
    \includegraphics[width=\textwidth]{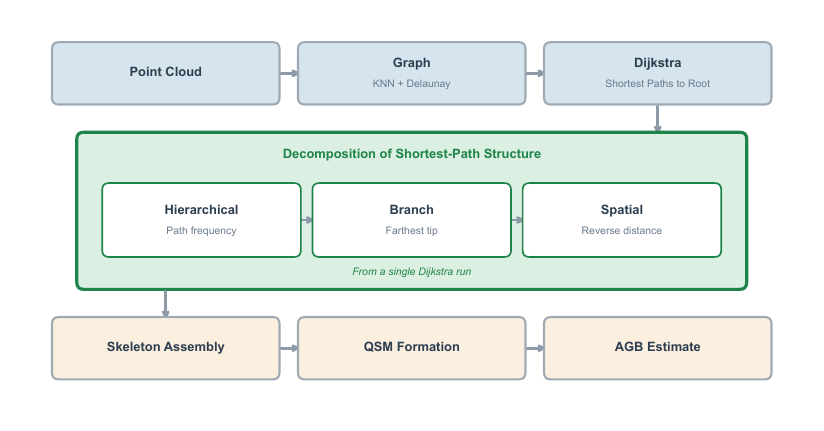}
    \caption{Pipeline overview. A hybrid KNN--Delaunay graph is constructed from the input point cloud. Dijkstra shortest-path analysis runs once. Three decompositions of the path structure recover the branching hierarchy. Branch diameters are modeled allometrically to form a QSM and compute AGB.}
    \label{fig:workflow}
\end{figure*}

\subsection{Graph Construction}
\label{sec:graph}

The input is a single-tree point cloud. The method proceeds in three stages, namely graph construction, path-based decomposition, and QSM assembly (Fig.~\ref{fig:workflow}). Algorithm~\ref{alg:pipeline} summarizes the data flow; the three decompositions operate on the output of a single Dijkstra run and are applied in sequence.

We construct the proximity graph $G = (V, E, W)$, whose formal properties are analyzed in Section~\ref{sec:theory}. Each node in $V$ corresponds to a point in the input cloud. Edges are built using $k$-nearest neighbors (KNN, $k=15$) and Delaunay triangulation, following the graph construction procedure of~\cite{WANG202167}, which includes local and global edge pruning to remove spurious connections and iterative Delaunay edge addition to ensure a single connected component. Each edge carries a weight $w_{ij}$ equal to the Euclidean distance between its endpoints. The root node $v_r$ is the lowest point along the vertical axis. For extraction, each undirected edge is replaced by two directed arcs with the asymmetric cost
\begin{equation}
\text{cost}(u \to v) = w_{uv}\bigl(1 + \beta\,\text{lat}^2(u)\bigr),
\label{eq:asym_cost}
\end{equation}
where $\text{lat}(u)$ is the horizontal distance from $u$ to the root axis and $\beta = 0.1$, a directed weighting in the spirit of the shortest-path routing of RayExtract~\cite{devereux2026rayextract}. Because the two directions carry different costs, a directed Dijkstra run computes the shortest paths from all nodes to $v_r$. The weighted search determines only the tree topology, namely the parent pointer of every node. For each node $v_i$ this yields its shortest path $\mathcal{P}_{\text{th}}(v_i)$, and the root distance $D_i$ is recomputed as the Euclidean path length along $\mathcal{P}_{\text{th}}(v_i)$, so that all measured distances remain in meters and the lateral penalty enters no downstream distance quantity. The set of all paths is denoted $\mathcal{P}$.

\subsection{Hierarchical Decomposition via Path Frequency}

Following the notation of Section~\ref{sec:theory}, the path frequency $\mathcal{F}_i$ of node $v_i$ is the number of shortest paths from all nodes to the root that traverse $v_i$.
\begin{equation}
\mathcal{F}_i = \bigl| \bigl\{ P \in \mathcal{P} \mid v_i \in P \bigr\} \bigr|.
\label{eq:Fi}
\end{equation}
By Theorem~1, $\mathcal{F}_i = |\operatorname{Sub}(v_i)|$ on the shortest-path tree extracted from $G$. Trunk nodes support large subtrees and accumulate high frequency; leaf-tip nodes support only themselves. By Corollary~1, $\mathcal{F}$ is monotonic along rootward paths on an ideal tree. In practice, the shortest-path tree extracted from the proximity graph $G$ may contain nodes that deviate from strict monotonicity due to measurement noise and redundant graph edges. We apply a propagation correction that restores the expected ordering.
\begin{equation}
\mathcal{F}_i' =
\begin{cases}
\displaystyle\max_{v_j \in \mathcal{N}(v_i), D_j < D_i} \mathcal{F}_j', & \text{if } \mathcal{F}_i < \max_{v_j \in \mathcal{N}(v_i), D_j < D_i} \mathcal{F}_j' \\[6pt]
\mathcal{F}_i, & \text{otherwise}
\end{cases}
\label{eq:Fi_correct}
\end{equation}
where $\mathcal{N}(v_i)$ is the neighbor set of $v_i$. Within each spatial cluster, the node with the highest $\mathcal{F}'$ becomes the skeleton node. Leaf-bearing extremities with intrinsically low $\mathcal{F}$ are implicitly suppressed.

\subsection{Branch Decomposition via Farthest Tip}

The farthest tip $\mathcal{T}$ identifies the terminal node with maximum root distance reachable via a given node.
\begin{equation}
\mathcal{T}_i = \arg\max_{t \in \mathcal{T}} \bigl( D_t \mid \exists P \in \mathcal{P}(v_i, t) \bigr),
\label{eq:Ti}
\end{equation}
where $\mathcal{T} = \{ v \in V \mid \text{out-degree}(v) = 0 \}$ is the set of terminal nodes. For nodes on the main trunk, which are reachable from multiple tips, we assign the tip with the largest root distance. We apply a propagation correction analogous to Eq.~(\ref{eq:Fi_correct}), enforcing monotonicity of $\mathcal{T}$ along rootward paths. If a node's assigned tip differs from that of a rootward neighbor, the node inherits the neighbor's tip provided the neighbor's corresponding root distance is larger. All nodes that share the same corrected farthest tip belong to the same branch. This partitions the point cloud into branch-level clusters $\mathcal{B}_t = \{v_i \in V \mid \mathcal{T}_i' = t\}$ using only path connectivity.

\subsection{Spatial Decomposition via Reverse Distance}

\begin{figure}[t]
    \centering
    \includegraphics[width=\ieeecolwidth]{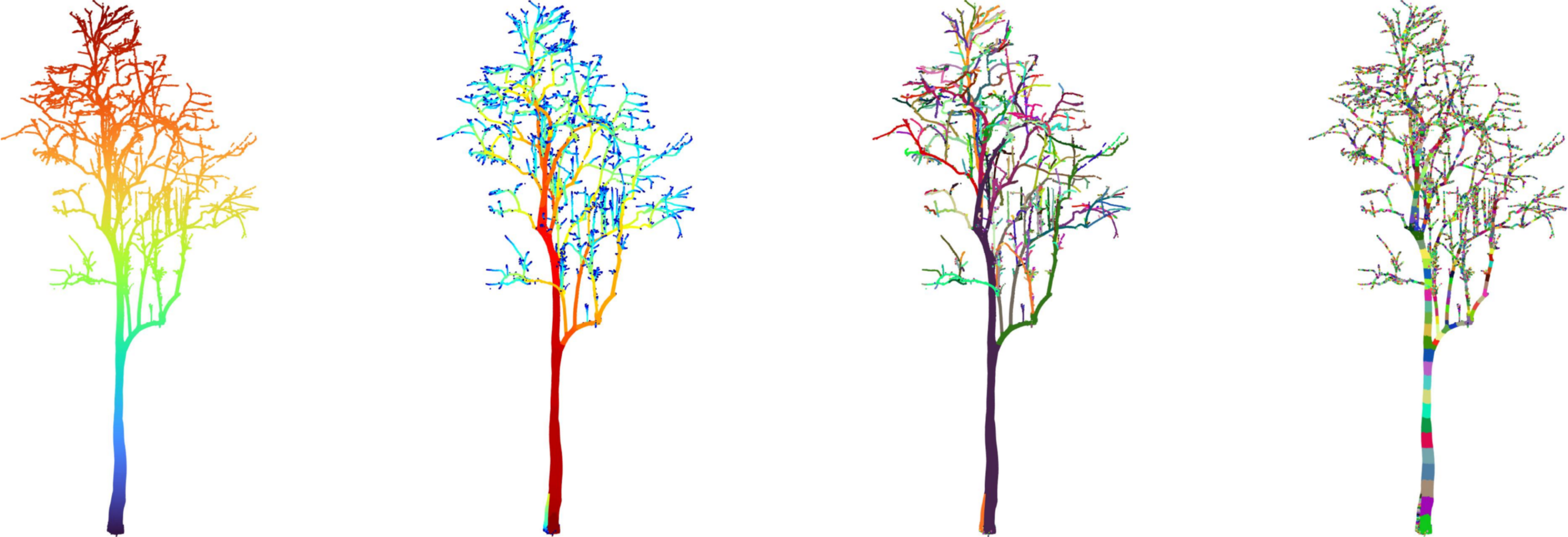}
    \caption{The three decompositions. From left to right, input point cloud, path frequency $\mathcal{F}$ in log scale, farthest tip $\mathcal{T}$ assignment, and adaptive clustering result.}
    \label{fig:clustering}
\end{figure}

\begin{figure}[t]
    \centering
    \includegraphics[width=0.6\ieeecolwidth]{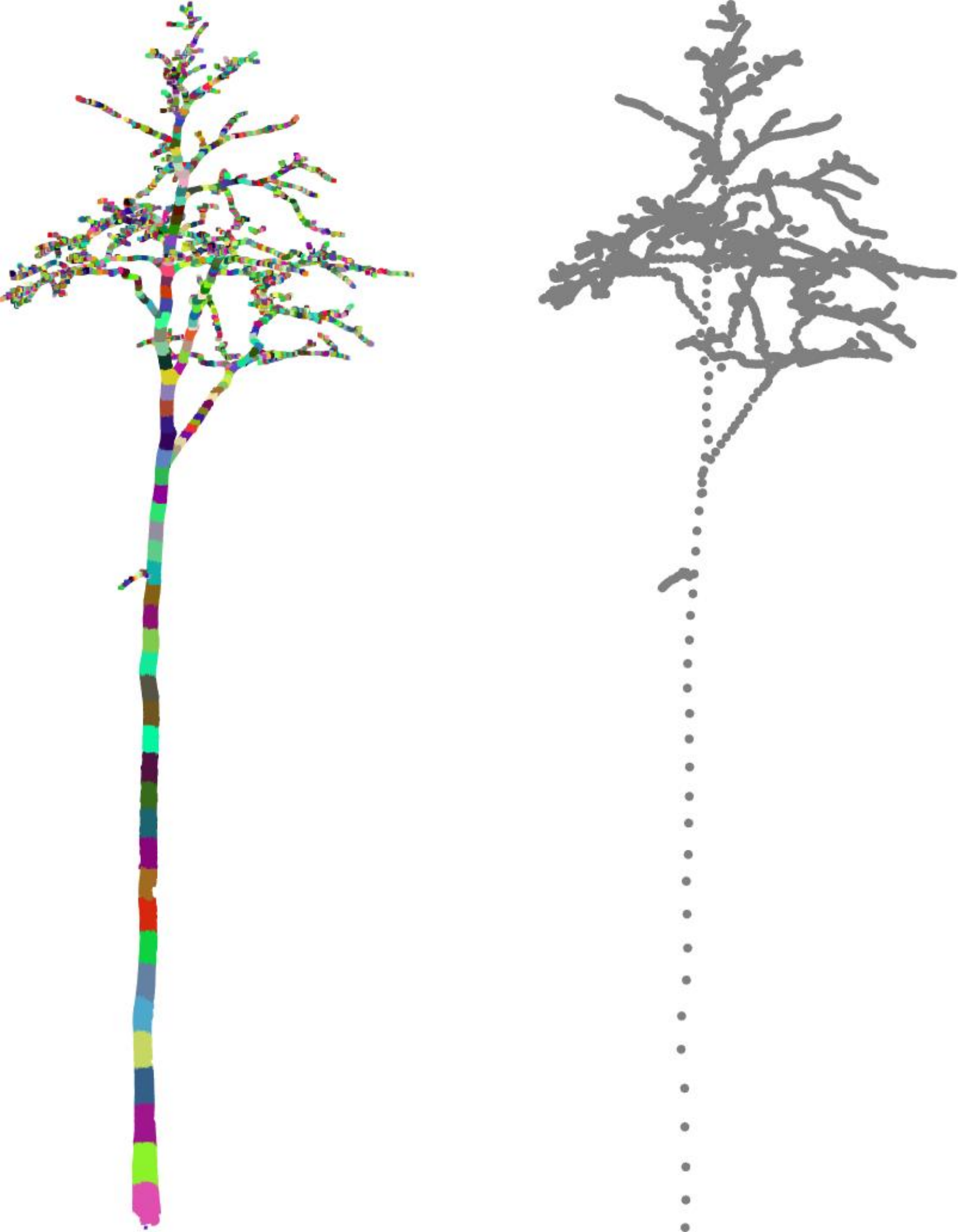}
    \caption{Adaptive clustering and resulting skeleton points. Detail is preserved in the crown; redundancy is avoided in the trunk.}
    \label{fig:skeletonpoints}
\end{figure}

Each branch cluster is segmented along its length. We define a reverse distance metric.
\begin{equation}
D_T(v_i) = D_{\mathcal{T}_i'} - D_i,
\label{eq:DT}
\end{equation}
The path length along the tree from node $v_i$ to its corrected farthest tip. An exponential step-size vector is defined on $D_T$.
\begin{align}
\mathbf{d}_s &= \frac{D_{T}^{\max} - D_{T}^{\min}}{a} \cdot \bigl( 10^{\mathbf{u}} - 1 \bigr) + D_{T}^{\min}, \\
\mathbf{u}   &= \operatorname{linspace}\!\bigl( 0,\ \log_{10}(a+1),\ 100 \bigr),
\label{eq:ds}
\end{align}
where $a = 20$. Since $D_T$ is small near branch tips and large near the trunk, this yields fine sampling in the crown and coarse sampling in the trunk. Under leaf-on conditions, mean-shift clustering with locally adaptive bandwidth preserves spatial resolution. Fig.~\ref{fig:clustering} illustrates the three decompositions on an example tree, and Fig.~\ref{fig:skeletonpoints} shows the resulting adaptive skeleton points.

\subsection{Adaptive Frequency Threshold}
\label{sec:adaptive}

A frequency threshold suppresses skeleton nodes in leaf-dominated regions where $\mathcal{F}$ is low. We determine this threshold adaptively per tree from the frequency growth pattern along each branch.

For each tip node, we walk rootward along the skeleton. Within a leaf cluster, frequency grows gradually as points from the same cluster accumulate. Where a lateral branch joins, the frequency jumps sharply. As a heuristic for detecting the leaf-to-branch transition, we use the first doubling of frequency along the path, which we term the \textit{elbow}. The median of elbow frequencies across all tips yields the per-tree adaptive threshold.
\begin{equation}
\mathcal{F}_{\text{adapt}} = \operatorname{median}_{t \in \text{tips}} \Bigl\{ \mathcal{F}\bigl(v_{\text{elbow}}(t)\bigr) \Bigr\},
\label{eq:adapt}
\end{equation}
where $v_{\text{elbow}}(t)$ is the first node on the path from tip $t$ toward the root at which the frequency exceeds twice the frequency of its child node.

\begin{figure}[t]
    \centering
    \includegraphics[width=\ieeecolwidth]{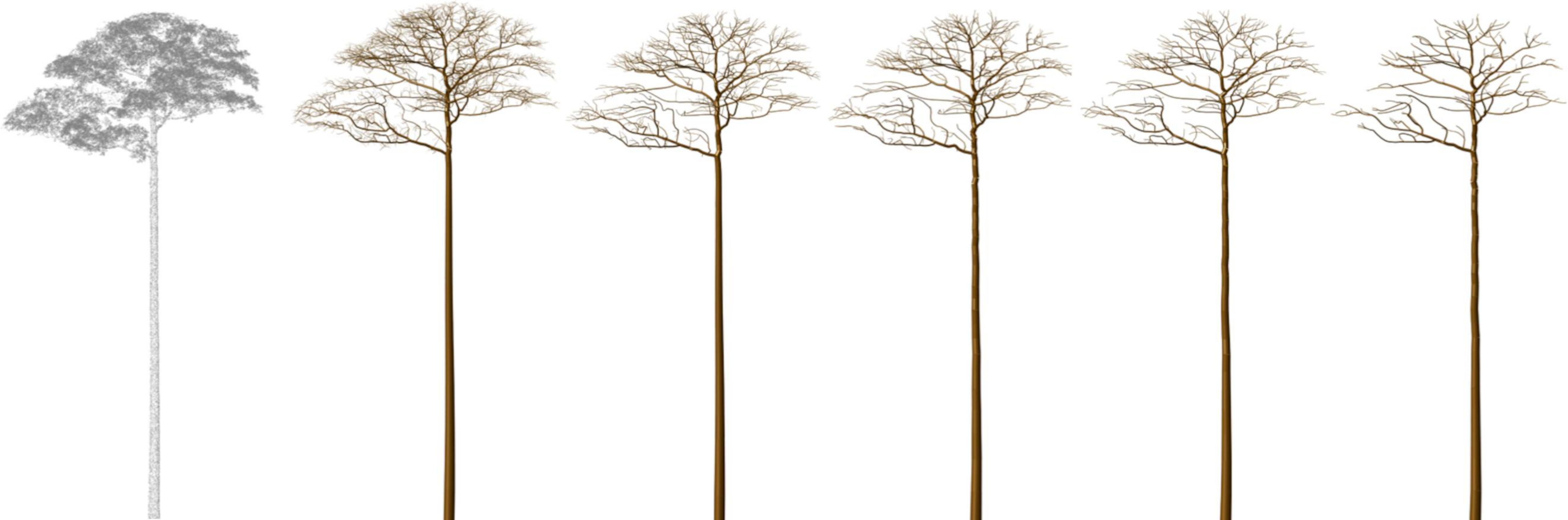}
    \caption{Effect of $\mathcal{F}$ thresholding on skeleton granularity. From left to right, the minimum frequency threshold increases, progressively suppressing leaf-bearing fine structures and retaining only the main branches.}
    \label{fig:granularity}
\end{figure}

The criterion has two properties worth noting. It is scale-invariant, since the decision depends only on the topological hierarchy encoded in frequency ratios, not on absolute magnitudes, tree height, or point density. The rule ``first doubling of frequency'' applies uniformly to all tips and all trees, requiring no user-specified threshold, window size, or distance constant. Fig.~\ref{fig:granularity} illustrates the effect of threshold choice on skeleton granularity.

The frequency threshold performs an implicit foliage suppression. Functionally, it separates leaf-associated nodes from branch-level nodes, but it does so using a topological quantity (path frequency) rather than local geometric features such as normals, linearity, or intensity~\cite{chen2025impact}. Because the separation emerges from the same path computation that builds the skeleton rather than from an independent preprocessing step, it avoids the classifier--reconstructor architecture whose error propagation is difficult to bound. Throughout this paper, ``foliage robustness'' refers to this mechanism, topology-driven hierarchical decomposition in place of explicit geometric leaf--wood classification.

\subsection{Model Assembly, QSM Formation, and AGB Computation}

Within each adaptive cluster, the node with the highest $\mathcal{F}$ is selected as a skeleton node and its position is adjusted to the $L_1$-median of its cluster~\cite{huang2013l1,feng2024l1}. Cluster-level connectivity is abstracted from node-level shortest paths without recomputation.

To form a QSM, branch diameters are estimated through allometric scaling. The root radius is obtained by circle-fitting at the trunk base. From the root, radii propagate outward. At branching nodes, radius is allocated to each child in proportion to its supported subtree length, following the pipe model~\cite{west1999general}.
\begin{equation}
R_c = R_p \cdot \left(\frac{l_c}{l_p}\right)^{\gamma},
\label{eq:branch_radius}
\end{equation}
where $l_c$ and $l_p$ are supported subtree lengths and $\gamma = 0.4$~\cite{chen2024towards}.

Cubic Hermite splines smooth each branch~\cite{he2018research}, and generalized cylinders define the surface geometry~\cite{hu2017efficient}. Tree volume is the sum of truncated-cone volumes over all skeleton edges. AGB is obtained by multiplying by species-specific wood density.

\subsection{Parameter Settings}

The skeleton extraction is governed by two free parameters, KNN size $k$ (default 15) and step-size control $a$ (default 20). The frequency threshold is determined adaptively per tree (Section~\ref{sec:adaptive}). Sensitivity to $k$ and $a$ is evaluated in Section~\ref{sec:sensitivity}.

\section{Experiments and Results}
\label{sec:results}

We evaluate model quality through AGB estimation accuracy against destructively measured reference values. Comparisons are against TreeQSM~\cite{Raumonen2013} and AdQSM~\cite{Fan2020}, both run with public implementations and default parameters. TreeQSM uses its automatic optimal model selection. To isolate the effect of leaf--wood separation, we also evaluate both QSMs on point clouds preprocessed with two leading separation algorithms, GBS~\cite{Tian2022} and LeWoS~\cite{wang2020lewos}, selected as the top performers in a recent benchmark~\cite{chen2025impact}. As a reference ceiling, we include leaf-off results where available; these represent the best-case performance of each QSM when foliage is absent. The following metrics quantify AGB estimation accuracy. MAPD, the mean absolute percentage deviation, reports the average absolute relative error; RMSE, the root mean square error, expresses the error magnitude in the original units; RMSE\%, the normalized root mean square error, expresses RMSE relative to the mean reference AGB; and Bias reports the mean signed percentage deviation.
\begin{align}
\text{MAPD} &= \frac{1}{n}\sum_{i=1}^{n}\frac{|\hat{y}_i - y_i|}{y_i} \times 100\%, \label{eq:mapd} \\[4pt]
\text{RMSE}  &= \sqrt{\frac{1}{n}\sum_{i=1}^{n}(\hat{y}_i - y_i)^2}, \label{eq:rmse} \\[4pt]
\text{RMSE\%} &= \frac{\text{RMSE}}{\bar{y}} \times 100\%, \label{eq:rmsepct} \\[4pt]
\text{Bias}  &= \frac{1}{n}\sum_{i=1}^{n}\frac{\hat{y}_i - y_i}{y_i} \times 100\%, \label{eq:bias}
\end{align}
where $\hat{y}_i$ and $y_i$ are estimated and reference AGB for tree $i$, $\bar{y}$ is the mean reference AGB, and $R^2$ is the squared Pearson correlation coefficient between estimated and reference AGB.

\subsection{Cameroon Leaf-On TLS}
\label{sec:cameroon}

\begin{figure*}[t]
    \centering
    \includegraphics[width=0.92\textwidth]{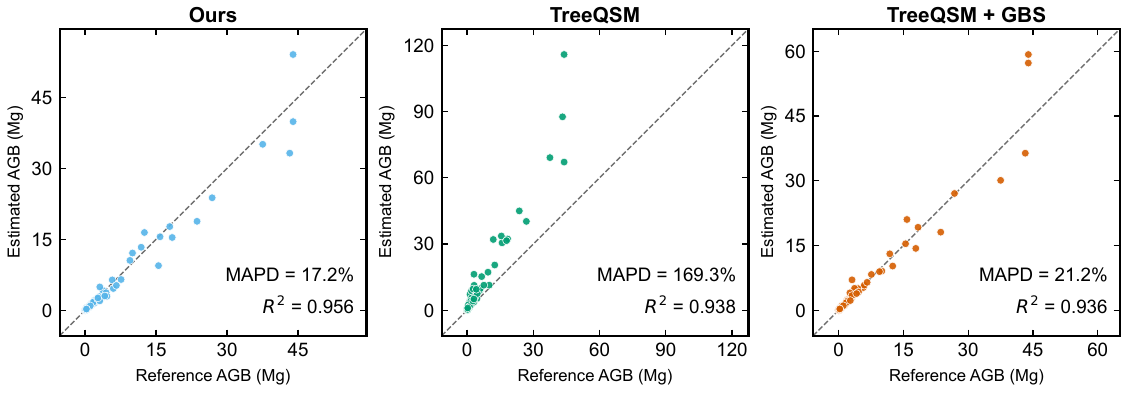}
    \caption{Estimated versus reference AGB on the Cameroon leaf-on dataset. Left, our method (MAPD 17.2\%). Center, TreeQSM (MAPD 169.3\%). Right, TreeQSM with GBS separation (MAPD 21.2\%). Dashed line indicates perfect agreement.}
    \label{fig:cameroon_scatter}
\end{figure*}

\begin{figure*}[t]
    \centering
    \includegraphics[width=0.88\textwidth]{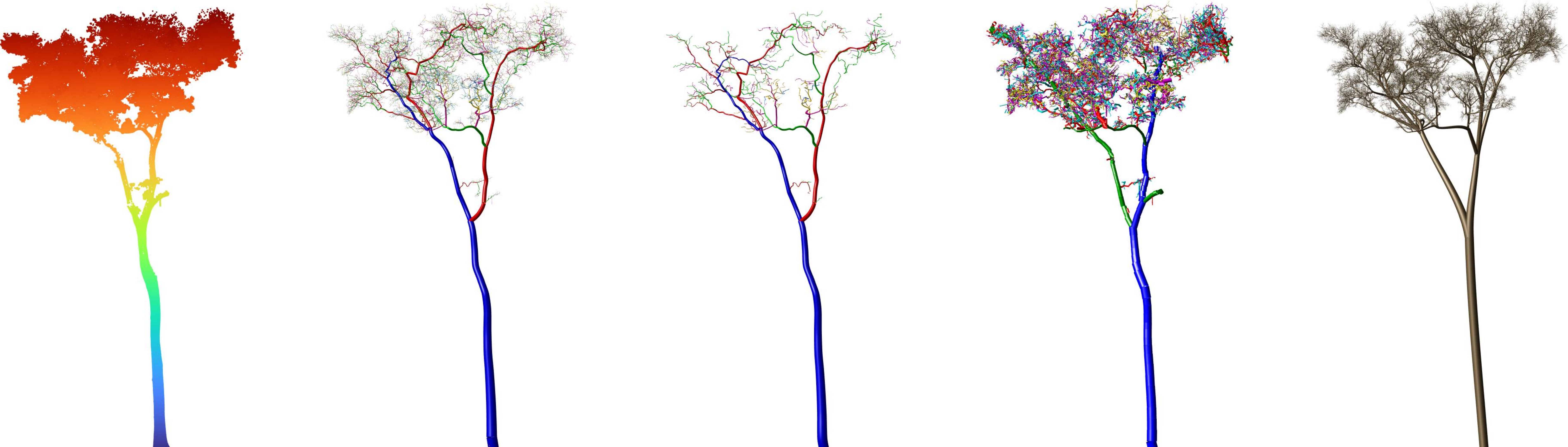}
    \caption{Reconstructed tree models under leaf-on conditions. Left to right, our method without $\mathcal{F}$ thresholding, with $\mathcal{F}$ thresholding, TreeQSM, AdQSM.}
    \label{fig:comparison}
\end{figure*}

\begin{table}[t]
\centering
\caption{AGB Estimation on Cameroon Dataset (61 trees, destructive AGB)}
\footnotesize
\setlength{\tabcolsep}{4pt}
\begin{tabular}{lccccc}
\toprule
Method & MAPD & RMSE & RMSE$\%$ & Bias & $R^2$ \\
       & (\%) & (Mg) & (\%)     & (\%) & \\
\midrule
\multicolumn{6}{c}{\textit{TreeQSM}} \\
\midrule
Leaf-off      & 11.0  & 1.81  & 26.5  & $-$1.5    & 0.974 \\
Leaf-on       & 169.3 & 13.67 & 200.3 & +169.3    & 0.938 \\
$+$ LeWoS     & 22.0  & 3.21  & 47.1  & +15.4     & 0.927 \\
$+$ GBS       & 21.2  & 3.18  & 46.5  & +11.8     & 0.936 \\
\midrule
\multicolumn{6}{c}{\textit{AdQSM}} \\
\midrule
Leaf-off      & 35.0  & 12.42 & 182.1 & $-$8.0    & 0.647 \\
Leaf-on       & 50.2  & 24.52 & 359.5 & +39.2     & 0.682 \\
$+$ LeWoS     & 22.9  & 2.68  & 40.3  & $-$5.1    & 0.950 \\
$+$ GBS       & 36.9  & 18.17 & 268.4 & $-$4.0    & 0.694 \\
\midrule
\multicolumn{6}{c}{\textit{Ours}} \\
\midrule
Leaf-off      & 20.6  & 3.27  & 47.9  & $-$3.8    & 0.932 \\
\textbf{Leaf-on} & \textbf{17.2} & \textbf{2.37} & \textbf{34.8} & \textbf{$-$0.6} & \textbf{0.956} \\
\bottomrule
\end{tabular}
\label{tab:cameroon}
\end{table}

Table~\ref{tab:cameroon} reports results on the Cameroon dataset. Under leaf-on conditions without separation, TreeQSM and AdQSM degrade severely, to 169.3\% and 50.2\% MAPD respectively, while our method achieves 17.2\%. Leaf--wood separation substantially recovers TreeQSM's accuracy, with GBS and LeWoS reducing MAPD to 21.2\% and 22.0\%. AdQSM benefits less consistently from separation, with LeWoS yielding 22.9\% MAPD while GBS reaches only 36.9\%. The RMSE values corroborate this ranking, with our method's 2.37~Mg the lowest of all leaf-on configurations, below the best separation-assisted results of 3.18~Mg for TreeQSM+GBS and 2.68~Mg for AdQSM+LeWoS, whereas unfiltered TreeQSM and AdQSM reach 13.67~Mg and 24.52~Mg. Our RMSE\% of 34.8\% is likewise the lowest of all leaf-on configurations, against 40.3\% for AdQSM+LeWoS. The leaf-off results establish the best-case performance of each QSM paradigm. TreeQSM achieves 11.0\% MAPD on clean data, confirming that cylinder fitting is highly accurate when foliage is absent. Our method's leaf-on MAPD of 17.2\% compares favorably to the leaf-off ceiling of 11.0\% without requiring leaf--wood separation. Our own leaf-off result reaches 20.6\% MAPD, slightly above the leaf-on value, a pattern analyzed in Section~\ref{sec:discussion}. Scatter plots of estimated versus reference AGB are shown in Fig.~\ref{fig:cameroon_scatter}, and reconstructed tree models are compared in Fig.~\ref{fig:comparison}.

The adaptive frequency threshold described in Section~\ref{sec:adaptive} produced per-tree values ranging from 18 to 56 with a median of 31 across the 61 trees and 15 species. This variation encodes genuine differences in branching architecture, as trees with deeper, more complex crowns produce higher thresholds because frequency doubles at more distal branch junctions. The threshold correlates positively with tree height, Pearson's $r = 0.41$, and with DBH, $r = 0.37$, confirming that larger trees possess more elaborate branching hierarchies that the elbow criterion captures without species-specific tuning.

TreeQSM's $R^2 = 0.938$, despite MAPD = 169.3\%, reveals that the leaf-on error is multiplicative rather than random. Cylinder fitting systematically overestimates volume by a factor roughly proportional to true AGB, consistent with foliage being fitted as additional woody cylinders whose aggregate volume scales with tree size. The high $R^2$ indicates a structural bias, not a methodological failure. TreeQSM produces predictably inflated output because every leaf cluster adds spurious volume.

\subsection{Low-Density ULS}
\label{sec:uls}

\begin{table}[t]
\centering
\caption{AGB Estimation on ULS Data by DBH Source, Germany (29 trees)}
\footnotesize
\setlength{\tabcolsep}{4pt}
\begin{tabular}{lccccc}
\toprule
DBH Source & MAPD & RMSE & RMSE$\%$ & Bias & $R^2$ \\
           & (\%) & (Mg) & (\%)     & (\%) & \\
\midrule
TLS-measured DBH   & 6.4  & 0.176 & 10.0  & $-$5.1 & 0.995 \\
Allometric ($H$--DBH) & 32.3 & 1.092 & 61.8  & $-$1.1 & 0.766 \\
ULS auto-DBH (9 of 29) & 17.9 & 0.881 & 24.4  & $-$16.1 & 0.943 \\
\bottomrule
\end{tabular}
\label{tab:uls}
\end{table}

\begin{figure}[t]
    \centering
    \includegraphics[width=\ieeecolwidth]{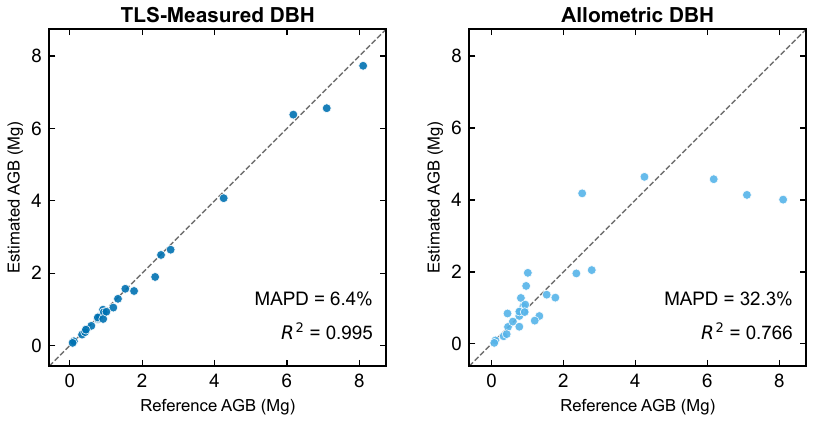}
    \caption{Estimated ULS-derived AGB versus TLS reference AGB on the Germany dataset. Left, TLS-measured DBH with 6.4\% MAPD. Right, allometric DBH from tree height with 32.3\% MAPD.}
    \label{fig:uls_scatter}
\end{figure}

\begin{figure}[t]
    \centering
    \includegraphics[width=0.85\ieeecolwidth]{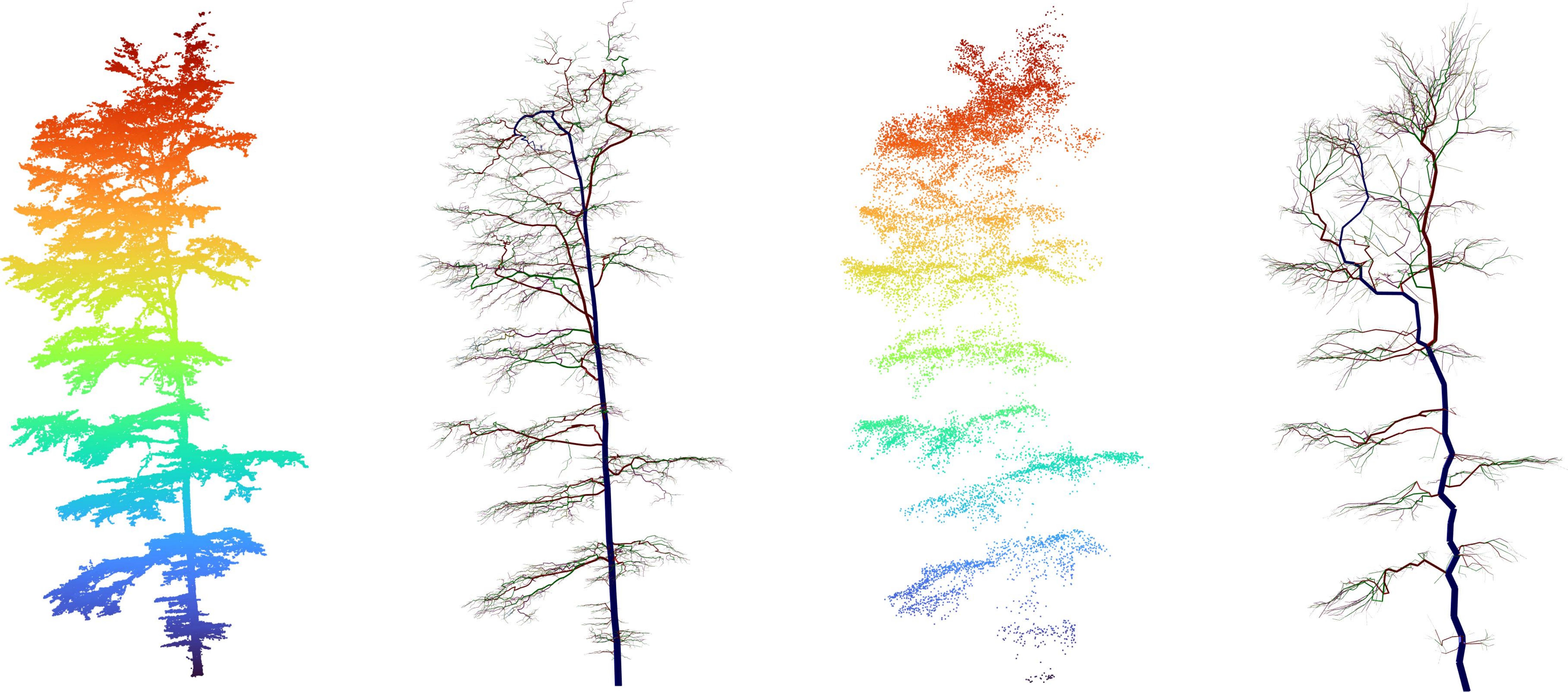}
    \caption{Reconstructed models from TLS (left) and ULS (right) data for the same tree. Skeleton topology is preserved despite lower ULS point density.}
    \label{fig:ulsmodelvis}
\end{figure}

Table~\ref{tab:uls} reports AGB estimation results on the Germany ULS dataset. The dataset tests the method under the combination of leaf-on foliage, low point density at roughly 100-fold reduction from TLS, and partial trunk coverage. All skeleton-extraction parameters remain identical to the Cameroon experiment. Direct ULS DBH estimation is a well-known difficulty, since overhead scanning often captures only partial trunk arcs and automated diameter fitting on such clouds is unreliable~\cite{brede2019nondestructive}. In our runs a valid automatic DBH estimate was obtained for only 9 of the 29 trees. On those 9 trees the method reaches 17.9\% MAPD with an $R^2$ of 0.943, and the systematic underestimation (Bias $-$16.1\%) is consistent with the automatically fitted diameters running below the TLS reference, confirming that when a usable diameter is available the topology extracted from sparse ULS data is sufficient to recover reference AGB. Allometric DBH inference from ULS-measured tree height provides a practical alternative, reaching 32.3\% MAPD. With TLS-measured DBH, ULS-derived AGB matches the reference closely, at 6.4\% MAPD, 0.176~Mg RMSE, and $R^2=0.995$, confirming that DBH estimation is the primary error source. We note that the 6.4\% MAPD on the 29 German trees should not be directly compared to the 17.2\% on the 61 Cameroon trees. The German dataset is architecturally simpler since it is temperate and species-poor, its reference AGB is TLS-derived rather than destructive, and the externally supplied DBH removes the largest single source of diameter error. Each factor makes the German comparison less stringent than the Cameroon one. The allometric DBH result at 32.3\% MAPD provides a more operationally realistic comparison point for ULS deployment without field measurements. Fig.~\ref{fig:uls_scatter} shows the scatter plots, and Fig.~\ref{fig:ulsmodelvis} compares reconstructed models from TLS and ULS data for the same tree.

\begin{table}[t]
\centering
\caption{Volume Fraction by Branch Order on Cameroon Dataset (61 trees, \%. Values are mean $\pm$ SE.)}
\footnotesize
\setlength{\tabcolsep}{3pt}
\begin{tabular}{l c c c c}
\toprule
Branch order & \shortstack{TreeQSM\\ leaf-off (ref.)} & \shortstack{TreeQSM + GBS\\ leaf-on} & \shortstack{Ours\\ leaf-off} & \shortstack{Ours\\ leaf-on} \\
\midrule
Order 1 & 75.4 $\pm$ 1.9 & 75.2 $\pm$ 1.7 & 80.7 $\pm$ 1.5 & \textbf{79.8 $\pm$ 1.4} \\
Order 2 & 14.4 $\pm$ 1.0 & 15.8 $\pm$ 1.0 & 12.5 $\pm$ 0.9 & \textbf{11.4 $\pm$ 0.8} \\
Order 3 & 7.5 $\pm$ 0.7 & 7.3 $\pm$ 0.7 & 5.0 $\pm$ 0.5 & \textbf{5.8 $\pm$ 0.4} \\
Order 4 & 2.7 $\pm$ 0.4 & 1.7 $\pm$ 0.2 & 1.5 $\pm$ 0.2 & \textbf{2.3 $\pm$ 0.2} \\
Order $5+$ & 0.48 $\pm$ 0.10 & 0.16 $\pm$ 0.03 & 0.32 $\pm$ 0.06 & \textbf{0.71 $\pm$ 0.09} \\
\bottomrule
\end{tabular}
\label{tab:volume_order}
\end{table}

\subsection{Structural Consistency Analysis}
\label{sec:topology}

\begin{figure*}[t]
    \centering
    \includegraphics[width=0.75\textwidth]{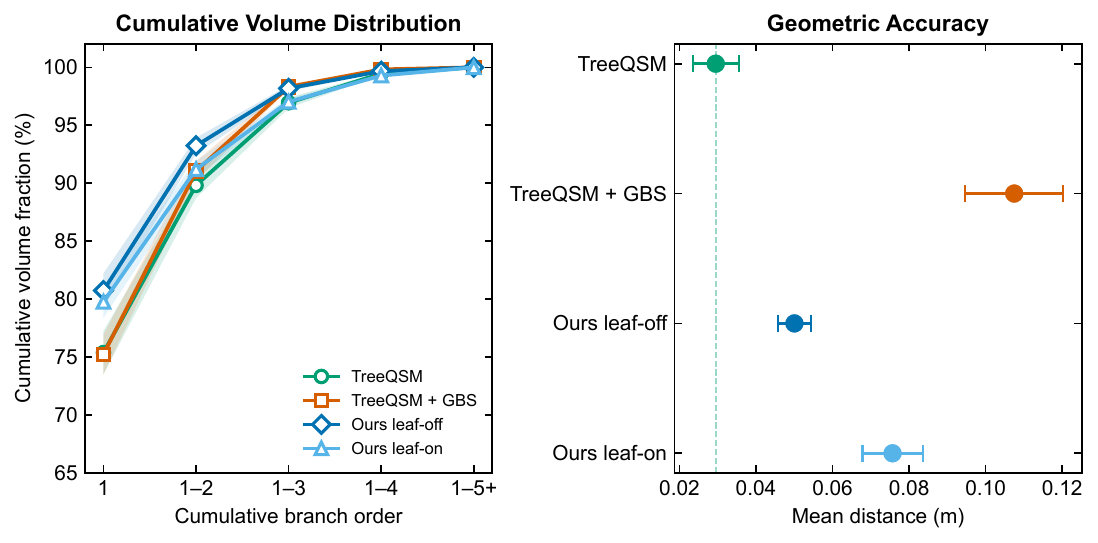}
    \caption{Structural consistency analysis on 61 Cameroon trees. Left, cumulative volume fraction by branch order, with shaded bands showing $\pm$1 SE. Right, mean point-to-model distance with error bars of $\pm$1 SE.}
    \label{fig:topology}
\end{figure*}

The AGB results in Sections~\ref{sec:cameroon} and \ref{sec:uls} demonstrate accurate biomass estimation, but aggregate AGB can mask compensating topological errors. We evaluate the structural correctness of the extracted skeleton directly on the 61 Cameroon trees through two complementary analyses, volume distribution by branch order (Table~\ref{tab:volume_order}) and point-to-model geometric accuracy (Table~\ref{tab:point2model}). Because branch-level ground truth is unavailable for mature tropical trees, the TreeQSM reconstruction from manually cleaned leaf-off scans is adopted as the best available structural reference rather than as ground truth. Throughout this section, $\pm$ values denote the standard error of the mean (SE) across the 61 trees. Figure~\ref{fig:topology} summarizes both analyses.

\begin{table}[t]
\centering
\caption{Point-to-Model Distance on Cameroon Dataset (61 trees, m. Values are mean $\pm$ SE.)}
\footnotesize
\begin{tabular}{l c c}
\toprule
Method & Mean $\pm$ SE & RMS $\pm$ SE \\
\midrule
TreeQSM (ref.) & 0.029 $\pm$ 0.006 & 0.073 $\pm$ 0.015 \\
TreeQSM + GBS & 0.107 $\pm$ 0.013 & 0.275 $\pm$ 0.026 \\
Ours leaf-off         & 0.050 $\pm$ 0.004 & 0.070 $\pm$ 0.006 \\
\textbf{Ours leaf-on} & \textbf{0.076 $\pm$ 0.008} & \textbf{0.108 $\pm$ 0.011} \\
\bottomrule
\end{tabular}
\label{tab:point2model}
\end{table}

\subsubsection{Volume Distribution by Branch Order}

Branch order represents the hierarchical organization of a tree rather than its geometry. Order 1 is the trunk, order 2 denotes first-order lateral branches, and higher orders correspond to progressively finer branching. A topology-preserving reconstruction should therefore reproduce not only the total biomass but also its distribution across hierarchical levels. Branch-order volume allocation thus provides an indirect assessment of topology recovery that is independent of the specific QSM algorithm used.

We compute per-tree volume fraction at each branch order for four configurations, namely TreeQSM leaf-off as the reference ceiling; TreeQSM + GBS leaf-on; our method leaf-off, with the frequency threshold set to 1 to retain the complete skeleton; and our method leaf-on, using the adaptive threshold under operational conditions. Branch orders are identified using a common graph traversal on each reconstructed skeleton.

Trunk volume deviates from the reference by at most 5.3 percentage points across all methods, confirming that the main axis is universally correct regardless of reconstruction strategy. The critical difference emerges at order $5+$. TreeQSM~+~GBS allocates only $0.16\%$ of volume to fine branches, one third of the reference value of $0.48\%$. GBS removes peripheral branch points that are topologically arterial but geometrically indistinguishable from leaves. Our method under leaf-on conditions retains $0.71\%$ at order $5+$, exceeding the TreeQSM leaf-off reference of $0.48\%$ and more than four times the GBS baseline. Path frequency correctly identifies these fine branches because they support subtrees, regardless of their local geometric appearance.

\subsubsection{Geometric Accuracy}

Since leaf-off wood points represent the observable woody surface with minimal foliage interference, the point-to-model distance provides a direct measure of geometric reconstruction fidelity. We compute the mean Euclidean distance from the leaf-off wood-only points to the reconstructed cylinder surface. The leaf-off points were manually cleaned in the original dataset. A structurally consistent model should pass near the true wood points; large distances indicate misaligned or missing cylinders.

The leaf-on to leaf-off degradation for our method is $0.026$\,m on the mean and $0.038$\,m on the RMS, meaning that foliage adds 2.6~cm of mean and 3.8~cm of RMS geometric error. TreeQSM~+~GBS produces an RMS distance of $0.275$\,m, 2.5 times worse than our leaf-on RMS of $0.108$\,m. Separation-induced point removal distorts the geometric surface more than our integrated topology approach.

\subsection{Parameter Sensitivity}
\label{sec:sensitivity}

\begin{figure}[t]
    \centering
    \includegraphics[width=\ieeecolwidth]{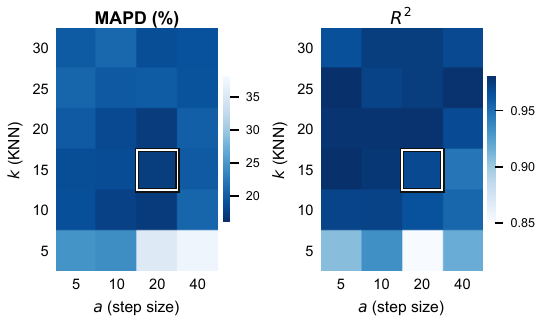}
    \caption{Parameter sensitivity on Cameroon leaf-on data. Left, MAPD (\%) across KNN size $k$ and step-size control $a$. Right, $R^2$ across $k$ and $a$. The default configuration ($k=15$, $a=20$) is outlined.}
    \label{fig:sensitivity}
\end{figure}

We evaluate sensitivity to the two free parameters, KNN size $k$ and step-size control $a$, over a grid of 24 configurations on the Cameroon dataset under both leaf conditions. The default configuration ($k=15$, $a=20$) ranks first on leaf-on MAPD, RMSE, and $R^2$, and its leaf-off MAPD of $20.6\%$ is the lowest among the configurations that lead on leaf-on accuracy. Fig.~\ref{fig:sensitivity} reports the leaf-on MAPD and $R^2$ surfaces. Excluding the $k=5$ row, whose sparse connectivity degrades fine-branch decomposition systematically (MAPD 24--37\%, RMSE\% 82--353\%, $R^2$ 0.62--0.93), the remaining 20 configurations span a narrow range, with MAPD from $17.0\%$ to $20.7\%$, RMSE\% from $31.5\%$ to $55.8\%$, and $R^2$ from 0.945 to 0.985, placing the default inside the plateau rather than at a fragile tuning point.

\section{Discussion}
\label{sec:discussion}

\begin{figure}[t]
    \centering
    \includegraphics[width=0.7\ieeecolwidth]{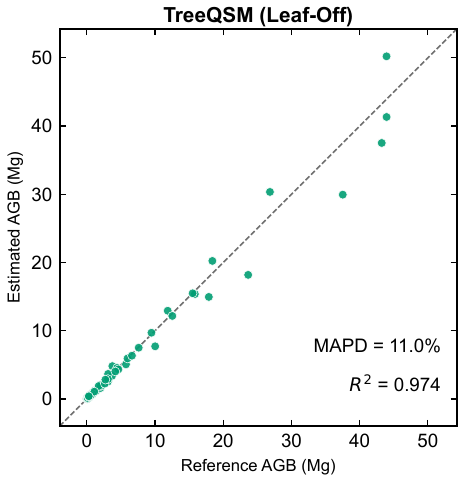}
    \caption{TreeQSM under leaf-off conditions (Cameroon, 61 trees). MAPD 11.0\%, representing the best-case ceiling for cylinder-fitting QSM when foliage is absent.}
    \label{fig:leafoff}
\end{figure}

\subsection{Foliage Robustness and the Separation Ceiling}

The Cameroon results establish a clear performance hierarchy, shown in Fig.~\ref{fig:leafoff} and Table~\ref{tab:cameroon}. TreeQSM achieves near-perfect accuracy on leaf-off data, with 11.0\% MAPD, confirming that cylinder fitting is highly effective when foliage is absent. Under leaf-on conditions, however, it collapses to 169.3\%, a 15-fold degradation. Leaf--wood separation recovers much of this loss. The best separation-assisted TreeQSM result reaches 21.2\% MAPD with GBS, a substantial improvement but still double the leaf-off baseline. Our method achieves 17.2\% without any separation preprocessing, reducing the leaf-on error by 9.8-fold relative to unfiltered TreeQSM and outperforming the best separation-assisted pipeline by 4.0\%.

This result challenges the implicit assumption that leaf--wood separation is a necessary preprocessing step for QSM-based AGB estimation. The best separation algorithm paired with the best QSM produces an error of 21.2\%; our method, which embeds foliage suppression in the reconstruction mechanism itself, reaches 17.2\%. The remaining gap to the leaf-off ideal of 11.0\% reflects the irreducible challenge of allometric diameter modeling versus direct cylinder measurement, not a failure to handle foliage.

Path-frequency decomposition sidesteps the need for a separate classification step because the frequency gap is topological rather than geometric. It persists regardless of leaf shape, point density, or sensor platform. The same parameters transfer from dense tropical TLS to sparse temperate ULS because branching topology, not local geometry, determines the arterial-to-peripheral ranking. The adaptive threshold in Section~\ref{sec:adaptive} operationalizes this property without per-dataset tuning.

As discussed in Section~\ref{sec:adaptive}, the frequency threshold performs an implicit foliage suppression operating on topological rather than geometric criteria. The structural consistency analysis in Section~\ref{sec:topology} provides direct evidence for this claim. Theorem~1 guarantees that path frequency encodes subtree cardinality on the shortest-path tree; fine peripheral branches, although geometrically similar to leaves, support subtrees and therefore accumulate path frequency above the leaf-level baseline. This is why our method preserves $0.71\%$ of volume at branch order $5+$ under leaf-on conditions while GBS-based separation reduces it to $0.16\%$ (Table~\ref{tab:volume_order}). GBS operates on local graph structure and cannot distinguish a thin branch supporting a subtree from a leaf cluster with similar connectivity. The point-to-model distance analysis confirms the result geometrically, since the RMS error of TreeQSM~+~GBS is $0.275$\,m, roughly 2.5 times larger than our leaf-on value of $0.108$\,m (Table~\ref{tab:point2model}), reflecting separation-induced surface distortion that our integrated topology approach avoids.

Our method under leaf-off conditions reaches $20.6\%$ MAPD, slightly above its leaf-on value of $17.2\%$, the reverse of the expectation that removing foliage eases reconstruction. Two observations from the on/off comparison point to the source of the shortfall. The leaf-off reference clouds are cleaned aggressively, which removes not only leaves but also many fine peripheral branches, and the leaf-off bias is more negative ($-$3.8\% against $-$0.6\% under leaf-on). Because the allometric radius propagation of Eq.~\ref{eq:branch_radius} allocates radii from the base over the supported subtree, the shorter supported subtree after aggressive cleaning offers a plausible account of the more negative bias, a mechanism that aggregate AGB cannot verify directly. On the largest tree, the complete-skeleton threshold retains base clutter and inflates the automatic diameter to 2.30~m against a reference of 1.80~m, an overestimate that the allometric model then carries into AGB. Cylinder-fitting baselines are comparatively insensitive to fine-branch removal, since fine branches contribute a small share of total volume, order $5+$ accounting for at most $0.71\%$, whereas our allometric coupling propagates the removal through the radius model. The leaf-off penalty therefore appears in AGB rather than in geometry, since the structural analysis (Section~\ref{sec:topology}) shows the leaf-off model fits the cleaned wood points more closely than the leaf-on model, the shortfall being precisely the fine-branch biomass that aggressive cleaning removed.

\subsection{Interaction Between Separation Algorithm and QSM Paradigm}

The Cameroon results in Table~\ref{tab:cameroon} reveal an asymmetry in how separation benefits each QSM paradigm. TreeQSM is acutely vulnerable to foliage, at 169.3\% MAPD, but recovers strongly when separation removes leaf points, reaching 21.2\% with GBS. AdQSM degrades less under foliage, at 50.2\%, but also benefits less from separation, with results varying from 22.9\% with LeWoS to 36.9\% with GBS. This interaction is consistent with the benchmark findings of~\cite{chen2025impact}, where separation quality depends on forest structure and errors propagate differently through cylinder-fitting and skeleton-based pipelines.

The divergent AdQSM results merit examination. GBS and LeWoS differ in how aggressively they remove peripheral points. LeWoS, with its recursive geometric segmentation, preserves more fine branch structure; GBS, relying on shortest-path connectivity patterns, removes more points from branch tips. AdQSM's MST-based reconstruction is sensitive to point density along branches, so when GBS thins branch extremities excessively, AdQSM either omits those branches or reconstructs them with incorrect topology, inflating volume error. TreeQSM's cylinder fitting is more tolerant of moderate thinning because it fits cylinders to whatever points remain. This interaction illustrates the architectural fragility of two-stage pipelines, where the separation algorithm is optimized independently of the downstream QSM and their interaction at the point-cloud level is untested by either. A method that integrates foliage suppression into the reconstruction mechanism avoids this interface altogether. Our method attains 17.2\% without any separation step, outperforming the best separation-assisted pipeline of 21.2\% (Table~\ref{tab:cameroon}).

\subsection{Topology-First vs.\ Geometry-First Reconstruction}

The results above point to a broader distinction between two reconstruction paradigms. Existing QSM methods recover topology as a by-product of cylinder fitting, since geometric primitives are fitted to local point neighborhoods and the skeleton emerges from connecting these primitives. Under this geometry-first approach, topology quality is bounded by the accuracy of local geometric operations, which are sensitive to point density, occlusion, and foliage-induced surface perturbations. The proposed framework inverts this order, recovering topology first from global graph connectivity and subsequently performing geometric modeling on the extracted skeleton. Because graph topology depends on connectivity patterns rather than local point geometry, it is considerably less sensitive to the factors that degrade geometry-first methods. This shift from geometry-first to topology-first reconstruction explains the method's consistent performance across foliage conditions, sensor platforms, and forest types.

\subsection{Cross-Platform Parameter Consistency}

A reconstruction principle based on graph topology should transfer across sensor platforms without modification, because the shortest-path structure depends on connectivity rather than absolute point density. The ULS results support this expectation. With independently provided DBH, the method achieves MAPD of 6.4\% on ULS data (Table~\ref{tab:uls}, Fig.~\ref{fig:uls_scatter}), a sensor platform for which most QSM methods were neither designed nor validated. The allometric DBH strategy, at 32.3\% MAPD, provides a practical alternative when field measurements are unavailable. It places the method within the range of published ULS-based AGB estimates while using no platform-specific tuning. Direct DBH estimation from ULS remains unreliable due to incomplete trunk coverage, a limitation shared by all QSM methods on this platform~\cite{brede2019nondestructive}.

\subsection{Limitations}

The method inherits limitations common to skeleton-based QSM approaches. It assumes single-trunk architecture. Branch diameters are modeled allometrically rather than fitted to data, which bounds accuracy below cylinder-fitting methods on clean input and ties AGB estimates to DBH quality. The adaptive threshold assumes a detectable frequency elbow. In extremely sparse point clouds where the first branch junction is poorly sampled, this assumption may weaken. The structural consistency analysis in Section~\ref{sec:topology} provides branch-level evaluation against a reference reconstruction from clean leaf-off data, but broader validation across additional forest types, particularly dense tropical forests with closed canopies, would establish wider generalizability. Trunk diameter estimation from sparse ULS remains an open challenge shared by all QSM methods on this platform.

\section{Conclusion}
\label{sec:conclusion}

This paper addressed the problem of foliage-robust AGB estimation from point clouds. The central finding is that recovering tree topology prior to geometric modeling provides an effective alternative to geometric leaf--wood classification for QSM reconstruction. We leveraged a topological property of rooted trees, namely that shortest-path traversal frequency equals subtree cardinality (Theorem~1), to extract branching hierarchy directly from graph connectivity. Three decompositions of the path structure, combined with an adaptive elbow threshold, recover the skeleton from a single Dijkstra run under identical parameter settings across all data sources. Branch diameters are modeled allometrically from the extracted topology; this design choice trades geometric precision for foliage robustness.

Experiments on 90 trees across tropical TLS and temperate ULS platforms, exclusively under leaf-on conditions, support three conclusions. First, topology-first reconstruction achieves MAPD of 17.2\% on dense tropical TLS data, outperforming the best separation-assisted pipeline (21.2\%). Second, structural consistency analysis confirms that foliage introduces 2.6~cm of mean geometric error (3.8~cm RMS) relative to leaf-off conditions, and that the hierarchical volume distribution across branch orders is preserved. Third, with externally provided DBH, the skeleton topology extracted from sparse ULS data recovers reference AGB (MAPD 6.4\%); the primary error source is DBH estimation, with allometric inference from tree height yielding MAPD 32.3\%. These results demonstrate that topology-first reconstruction, by replacing geometric leaf--wood classification with topology-driven foliage suppression, enables robust QSM estimation across sensor platforms and forest types.

\section*{Code Availability}
The source code implementing the proposed method is publicly available at \url{https://github.com/dwang520/GraphSkeleton}.

\end{document}